\documentclass{article}

\usepackage{microtype}
\usepackage{graphicx}
\usepackage{subcaption}
\usepackage{booktabs} 

\usepackage{booktabs}
\usepackage{tabularx}
\usepackage{multirow}

\usepackage{hyperref}

\usepackage[preprint]{icml2026}

\usepackage{amsmath}
\usepackage{amssymb}
\usepackage{mathtools}
\usepackage{amsthm}

\usepackage[capitalize,noabbrev]{cleveref}

\theoremstyle{plain}

\theoremstyle{definition}

\theoremstyle{remark}

\usepackage[textsize=tiny]{todonotes}

\icmltitlerunning{Future-as-Label: Scalable Supervision from Real-World Outcomes}

\begin{document}

\twocolumn[
  \icmltitle{Future-as-Label: Scalable Supervision from Real-World Outcomes}




  \begin{icmlauthorlist}
    \icmlauthor{Benjamin Turtel}{lrl}
    \icmlauthor{Paul Wilczewski}{lrl}
    \icmlauthor{Danny Franklin}{lrl}
    \icmlauthor{Kris Skothiem}{lrl}
  \end{icmlauthorlist}

  \icmlaffiliation{lrl}{Lightning Rod Labs}

  \icmlcorrespondingauthor{Benjamin Turtel}{ben@lightningrod.ai}

  \icmlkeywords{Machine Learning, ICML}

  \vskip 0.3in
]



\printAffiliationsAndNotice{}  

\begin{abstract}
Time creates free supervision: forecasts about real-world events resolve to verifiable outcomes. The passage of time provides labels that require no annotation. To exploit this structure, we extend reinforcement learning with verifiable rewards to real-world prediction over time. We train language models to make probabilistic forecasts from causally masked information, using proper scoring rules as the reward function once events resolve. Learning is driven entirely by realized outcomes, enabling scalable outcome-based supervision in open-world prediction. On real-world forecasting benchmarks, Qwen3-32B trained using Foresight Learning improves Brier score by 27\% and halves calibration error relative to its pretrained baseline, and outperforms Qwen3-235B on both constructed future-event prediction tasks and the Metaculus benchmark despite a 7× parameter disadvantage.
\end{abstract}

\noindent\textbf{Training data:}
\href{https://huggingface.co/datasets/LightningRodLabs/future-as-label-paper-training-dataset}
{Hugging Face dataset}

\noindent\textbf{Model weights:} 
\href{https://huggingface.co/LightningRodLabs/future-as-label-paper-step160}
{Hugging Face model repo}

\noindent\textbf{Data generation:} \url{https://lightningrod.ai/}

\section{Introduction}

Reinforcement learning with verifiable rewards has recently emerged as
an effective approach for improving language models in domains such as
mathematics, code generation, and formal reasoning, where correctness
can be checked automatically. By replacing human annotation with
deterministic reward functions, these methods scale efficiently and
yield strong empirical gains. However, their applicability depends on
the availability of immediate, closed-form verification, restricting
them to tasks where correctness can be resolved at training time. As a
result, despite their success, existing RLVR approaches remain confined
to a narrow class of problems defined by readily available,
task-specific reward signals.

In contrast, many real-world processes evolve over time and resolve to
objective outcomes that are independent of the model. These outcomes,
such as the conclusion of an election or the decision in a court case,
are publicly observable and verifiable after the fact. This temporal
structure induces a natural asymmetry between the information available
at prediction time and the information revealed at resolution, creating
a setting in which predictions can be evaluated retrospectively without
relying on contemporaneous labels.

Our goal is to translate this temporal structure into a scalable
learning framework for language models. Building on prior work on
Foresight Learning \cite{Turtel25},
we formalize learning from real-world temporal streams by grounding
supervision in event resolution. The key constraint is causal: at
prediction time t, the model is restricted to information available up
to t, while evaluation is deferred until the corresponding outcome is
realized. This formulation extends reinforcement learning with
verifiable rewards beyond closed-world tasks with immediate feedback to
settings where correctness is determined only after external, real-world
resolution.

We adopt a reward-based objective that frames prediction as a stochastic
decision evaluated retrospectively after outcome resolution. In contrast
to supervised fine-tuning, which fits fixed targets, Foresight Learning
optimizes over sampled reasoning trajectories using only outcome-based
rewards, without intermediate annotations or task-specific labels. This
perspective emphasizes calibration and decision quality rather than
target matching. While we focus on binary outcomes for clarity, the
formulation generalizes naturally to richer outcome spaces, such as
continuous, multi-class, and free-text outcomes.

In this work, we formalize learning from temporally resolved real-world
events as an extension of reinforcement learning with verifiable
rewards, introduce an annotation-free algorithm for learning from
delayed, outcome-based supervision, and show that this approach yields
substantial improvements in calibration and predictive accuracy over
strong pretrained baselines.

\section{Related Work}

\subsection{Reinforcement learning with verifiable rewards}

Reinforcement learning with verifiable rewards (RLVR) has demonstrated
strong results in domains with immediate, algorithmically checkable
feedback, such as mathematics and programming \cite{Wen25} \cite{Su25}. 
These settings typically involve short horizons and tightly scoped
environments, which simplify credit assignment and reward attribution.
Foresight Learning extends this paradigm to settings where outcomes
resolve only after substantial temporal delay and outside the model's
control.

\subsection{LLM-based forecasting and static supervision}

Recent work applies large language models to forecasting real-world
events using prompting, retrieval, ensembling, and supervised
fine-tuning over historical questions \cite{Halawi24}.
In particular, Halawi et al. generate multiple candidate
reasoning--prediction pairs and then use realized outcomes (via Brier
score) to select high-performing outputs for supervised fine-tuning.
While outcome information is therefore used for offline data curation,
this learning remains non-interactive. Foresight Learning differs by
incorporating outcome resolution directly into the training loop as
reinforcement signals.

\subsection{Model-based judges and endogenous rewards}

Another line of work uses language models as evaluators or judges to
provide scalable feedback in settings where objective reward functions
are unavailable \cite{Liu25a}.
Such approaches enable efficient supervision, but the resulting rewards
are model-mediated and derived from re-evaluating the same information
available to the predictor, which can propagate the biases and
limitations of the judge model.

Foresight Learning differs in the source of supervision rather than the
evaluation mechanism itself. While language models may assist in outcome
resolution (e.g., assessing free-text evidence), the resolver has access
to information that is causally unavailable at prediction time. Rewards
are therefore grounded in externally resolved outcomes rather than
alternative interpretations of the same input, ensuring that supervision
reflects genuinely new evidence revealed over time.

\section{Method}

We consider settings where supervision is provided by the eventual
resolution of events rather than contemporaneous labels. At prediction
time t, the model observes only information available up to that cutoff
and predicts whether an event will occur by a later time
s \textgreater t. Although training is performed on events whose outcomes
are already known, inputs are causally filtered to exclude post-t
information, and rewards are computed solely from outcome resolution at
time s, preserving the temporal asymmetry of prediction by construction.

\subsection{Learning formulation}

Each episode corresponds to a single future-event prediction.

\textbf{Predictor and resolver roles.\\
}Foresight Learning decomposes learning from temporal streams into two
roles with asymmetric information access:

\begin{itemize}
\item
  The \textbf{predictor} is the language model being trained. At time t,
  it observes a temporally masked information state and produces a
  probabilistic prediction about a future event.
\item
  The \textbf{resolver} is an external, fixed process that determines
  the realized outcome once the event resolves at time \emph{s
  \textgreater{} t}. The resolver is implemented using a pretrained,
  frozen language model that is not trained, updated or influenced by
  the learning process. The resolver has access to post-t information
  unavailable to the predictor and is used solely to resolve outcomes,
  not score or rank predictions.
\end{itemize}

The predictor and resolver are strictly separated: the predictor never
observes resolution information, and the resolver does not observe model
outputs or training dynamics. Learning is driven by the
\textbf{information gap} between the predictor's masked view at time t
and the resolver's unmasked view at time s.

\textbf{State.\\
}The state consists of all information available up to time \emph{t},
including relevant dated text and a natural-language specification of an
event guaranteed to resolve by time \emph{s \textgreater{} t}. The
predictor operates under a masked information state, with all post-t
information causally excluded by construction.

\textbf{Action.\\
}Conditioned on the state, the policy samples an internal reasoning
trajectory terminates in a probabilistic prediction \( p \in (0,1) \), represented
as a scalar value rather than a generated token. Only this numeric
probability is exposed to the environment.. Formally, the action is the
emitted probability; the trajectory is an internal stochastic
computation optimized via policy gradients.

\textbf{Reward.\\
}Once the event resolves, a terminal reward is assigned using the log
score:

\[
\text{Reward} = y \cdot \log(p) + (1 - y) \cdot \log(1 - p)
\]

where \( y \in \{0,1\} \) is the realized outcome. This strictly proper
scoring rule incentivizes calibrated probabilistic predictions and
provides a continuous learning signal under uncertainty.

Outcome determination is performed by a separate resolver that observes
the unmasked future. The resolver has access to post-t sources
unavailable to the predictor and is used solely to verify whether the
event occurred. Each episode terminates after outcome resolution; there
are no intermediate rewards.

Although the terminal reward takes the form of a proper scoring rule,
this learning setup is not simply supervised likelihood training. In
expectation, optimizing this reward corresponds to maximizing the
log-likelihood of realized outcomes conditioned on the information
available at prediction time. However, the learning problem is
structured differently: the predictor acts under a causally masked
information state without access to outcomes, and training optimizes a
stochastic policy over reasoning trajectories whose quality is evaluated
only after outcome resolution. Credit assignment is performed via policy
gradients on sampled trajectories rather than by directly
differentiating a likelihood objective, preserving the
decision-theoretic structure of acting under asymmetric information.

This formulation treats prediction as a stochastic decision evaluated
retrospectively after outcome resolution, distinguishing it from
supervised likelihood training even though the reward takes the form of
a proper scoring rule.

\subsection{Objective and optimization}

The objective is to maximize expected terminal reward under
outcome-based supervision. In this regime, single-sample policy
gradients exhibit high variance due to sparse terminal feedback and
intrinsic uncertainty in event outcomes. To address this, we optimize
using \textbf{Group Relative Policy Optimization (GRPO)}, as formulated
by \cite{Liu25b}.

For each state, the policy samples a group of \emph{K} trajectories,
each producing a probabilistic prediction. After outcome resolution, a
reward is computed for each trajectory. We define a group-relative
advantage by subtracting the mean reward within the group:

\[
\text{Advantage}(\tau_i) = \text{Reward}(\tau_i)
- \frac{1}{K} \sum_{j} \text{Reward}(\tau_j)
\]

Policy updates maximize the expected advantage-weighted log probability
of each trajectory under the current policy.

By comparing trajectories generated under identical pre-t information,
GRPO reduces variance from outcome noise and stabilizes learning when
supervision is provided only through terminal outcomes. Gradients are
applied to all tokens in each trajectory, enabling credit assignment
across extended reasoning processes even though feedback is available
only at the final step.

\subsection{Training protocol}

We explicitly enforce a causal information constraint by applying a
temporal information mask to the input stream. For each prediction time
t, the predictor is restricted to observing only information timestamped
at or before t, even though training is performed offline. All training
events resolve strictly after the pretrained model's knowledge cutoff,
ensuring that realized outcomes cannot be encoded in the model's
parametric memory. All post-t information - including sources required
to determine the outcome - is withheld during prediction and policy
optimization. Outcome verification is performed by a separate resolver
with access to the unmasked stream. This preserves a strict causal
separation between observation, action, and verification throughout
training.

\begin{figure}[ht]
  \vskip 0.2in
  \begin{center}
    \centerline{\includegraphics[width=\columnwidth]{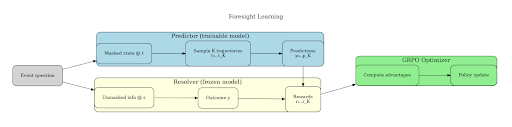}}
    \caption{
      Overview of Foresight Learning
    }
  \end{center}
\end{figure}

\section{Experimental Setup}

\subsection{Dataset construction}

We construct a future-event prediction dataset designed to preserve a
strict temporal separation between prediction and verification. The
pre-cutoff information state consists of an English-language news corpus
aggregated from publicly accessible outlets (e.g., international
newspapers, wire services, and financial news sites). Articles are
timestamped using publisher-provided publication times, normalized to
UTC.

For each example, we freeze the news corpus at a cutoff time t and
generate a binary question about an event expected to resolve strictly
after that cutoff, using only information available prior to t. The
cutoff is defined with respect to publisher-reported publication
timestamps; articles with missing or ambiguous timestamps are excluded
to prevent temporal leakage. Generated events span multiple domains,
including politics, economics, and corporate actions.

To prevent information leakage, model inputs are constructed exclusively
from sources published at or before time t. Outcome verification uses
independent post-t sources that are not included in the model's input
context. Event outcomes are resolved automatically by a separate, frozen
large language model (Gemini-2.5-Flash) with access to a broader pool of
post-cutoff news and archival sources, and used solely to determine
whether an event occurred. The resolver does not observe model outputs
or training dynamics; as a result, resolution errors introduce noise but
do not induce endogenous reward signals. Examples that cannot be
resolved with high confidence are discarded. Each event is assigned a
resolution time s, defined as the earliest dated source supporting the
resolved outcome.

All questions and outcomes are generated prior to training, enabling
fully offline optimization while preserving the temporal and causal
structure of real-world prediction.

\subsubsection{Dataset Statistics and Splits}

The full dataset contains 5,620 binary prediction examples. Of these,
5,120 examples are used for training, and 500 examples are held out as a
temporally disjoint test set constructed using the same event-generation
procedure. In addition, we evaluate on a second, independent test set
consisting of 293 human-written forecasting questions from Metaculus,
which are never used during training or data construction.

We intentionally do not construct a validation set: models are trained
on all available pre-test data using a fixed training procedure, without
early stopping or model selection based on held-out examples. Training
data consists of predictions made as of July 1, 2024 through January 30,
2025. Both test sets consist exclusively of predictions made on or after
February 1, 2025, ensuring strict temporal separation between training
and evaluation.

\subsubsection{Task characteristics}

Prediction horizons range from days to several weeks, allowing learning
across varying outcome horizons. Although outcomes are discrete,
supervision and evaluation are based on continuous probabilistic scores,
enabling analysis of accuracy and calibration under increasing temporal
uncertainty.

\subsection{Models and training}

We fine-tune a Qwen3-32B language model with explicit reasoning enabled.
Conditioned on an information state, the model generates a reasoning
trajectory that terminates in a probabilistic prediction expressed
explicitly at the end of the output. Parsed probabilities are
constrained to the interval {[}0.001, 0.999{]} for numerical stability.

Training is performed using GRPO. For each event, the model samples four
independent trajectories, each producing a probabilistic prediction.
After outcome resolution, a log-score reward is computed for each
trajectory, and relative advantages are obtained by subtracting the
per-group mean reward. Policy updates increase the relative likelihood
of higher-reward trajectories. Training uses batches of 32 events, with
prediction horizons mixed within each batch.

\subsubsection{Baselines}

We compare Foresight Learning to baselines that operate under identical
temporal constraints and produce probabilistic predictions in the same
output format, isolating the effect of learning.

\textbf{Prompted forecasting:} the base Qwen3-32B and Qwen-235B models
are prompted to produce probabilistic predictions without task-specific
fine-tuning. This baseline measures forecasting performance without
learning from outcome resolution.

\textbf{Ensembling:} multiple independent predictions are generated per
event and averaged to assess gains from sampling and aggregation without
parameter updates. This control tests whether improvements can be
explained by variance reduction alone.

\subsubsection{Evaluation metrics}

We evaluate models based on the quality of probabilistic predictions. We
report the \textbf{log score} used for training, the \textbf{Brier
score}, which measures squared error between predicted probabilities and
outcomes, and \textbf{calibration}, assessed via expected calibration
error (ECE) over 10 discretized probability bins measuring empirical
outcome frequencies as a function of predicted confidence.

\section{Results}

We evaluate Foresight Learning on two held-out test sets: (i) a
synthetic future-event benchmark of 500 questions constructed under
strict temporal controls, and (ii) an external benchmark consisting of
293 binary forecasting questions from Metaculus. Performance is
evaluated using proper scoring rules and calibration metrics.

\textbf{Table 1} compares four inference regimes: (i) Qwen3-32B prompted
for a single forecast, (ii) Qwen3-32B prompted for seven independent
forecasts with the median taken as the final prediction, (iii)
Qwen3-235B prompted for a single forecast, and (iv) the
Foresight-trained model prompted once. Repeated prompting and median
aggregation provide modest improvements over single-sample prompting but
do not match the gains from training on resolved outcomes. Notably, the
Foresight-trained 32B model outperforms both the ensemble-style baseline
and the substantially larger 235B model across all metrics, indicating
that the improvements stem from the training objective rather than
increased sampling or model scale.

Performance gains persist on the Metaculus benchmark, which consists of
independently authored questions outside the synthetic benchmark
distribution. One possible contributing factor is that Metaculus
questions often concern higher-salience events with broader public
coverage, providing richer information at prediction time. While this
hypothesis requires further study, the results indicate that learning
from externally resolved outcomes generalizes beyond the specific data
construction process used for training.

Taken together, these results support the central premise of Foresight
Learning: incorporating outcome resolution directly into the training
objective yields more accurate and better-calibrated probabilistic
forecasts than prompting or sampling-based baselines alone, even when
compared to substantially larger pretrained models.

\begin{table}[t]
  \caption{Forecasting performance on synthetic and real-world
benchmarks}
  \label{tab:calib}
  \begin{center}
  \begin{small}
  \setlength{\tabcolsep}{4pt}
  \renewcommand{\arraystretch}{1.05}

  \begin{tabularx}{\columnwidth}{Xrrr}
    \toprule
    \textbf{Model} & \textbf{Log $\uparrow$} & \textbf{Brier $\downarrow$} & \textbf{ECE $\downarrow$} \\
    \midrule

    \multicolumn{4}{l}{\textbf{Metaculus}} \\
    Qwen3-32B                   & -0.7210 & 0.2472 & 0.2175 \\
    Qwen3-32B Ensemble          & -0.7000 & 0.2390 & 0.2289 \\
    \textbf{Qwen3-32B-RL (160)} & \textbf{-0.5738} & \textbf{0.1793} & \textbf{0.1042} \\
    Qwen3-235B                  & -0.6828 & 0.2111 & 0.1905 \\
    \addlinespace

    \multicolumn{4}{l}{\textbf{Synthetic future-events}} \\
    Qwen3-32B                   & -0.7166 & 0.2432 & 0.1732 \\
    Qwen3-32B Ensemble          & -0.7045 & 0.2481 & 0.1864 \\
    \textbf{Qwen3-32B-RL (160)} & \textbf{-0.5978} & \textbf{0.1979} & \textbf{0.0598} \\
    Qwen3-235B                  & -0.7138 & 0.2260 & 0.1695 \\
    \bottomrule
  \end{tabularx}

  \end{small}
  \end{center}
  \vskip -0.1in
\end{table}

\begin{figure}[ht]
  \vskip 0.2in
  \begin{center}
    \centerline{\includegraphics[width=\columnwidth]{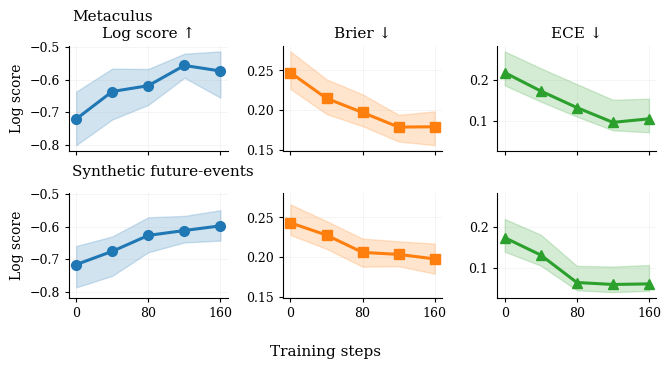}}
    \caption{Model calibration and accuracy metrics versus
training steps on Metaculus (top) and synthetic future-events (bottom).
Shaded regions show 95\% bootstrap confidence intervals. Metrics are log
score (↑), Brier score (↓), and expected calibration error (ECE; ↓).
Performance improves monotonically with training.
    }
  \end{center}
\end{figure}

\begin{figure}[ht]
  \vskip 0.2in
  \begin{center}
    \centerline{\includegraphics[width=\columnwidth]{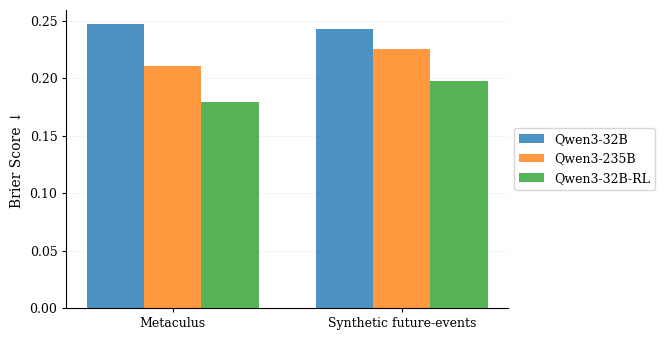}}
    \caption{Brier scores (↓) for different models on Metaculus
and synthetic future-events benchmarks. Foresight Learning consistently
outperforms both the base and larger pretrained baselines.
    }
  \end{center}
\end{figure}

\section{Discussion and Conclusion}

This work studies a supervision regime in which feedback is provided by
the eventual resolution of real-world events rather than contemporaneous
labels or proxy objectives. By optimizing probabilistic predictions
retrospectively using proper scoring rules, Foresight Learning aligns
training with the temporal and causal structure of forecasting under
uncertainty.

Empirically, learning from outcome resolution improves probabilistic
forecasting performance relative to a strong pretrained baseline, with
consistent gains in accuracy and calibration on both synthetic
future-event datasets and the independently authored Metaculus
benchmark. Notably, Foresight Learning materially outperforms a
substantially larger same-generation model on real-world forecasting
tasks.

A key benefit of outcome-based supervision is improved calibration.
Because rewards are assigned only after outcomes resolve, overconfident
incorrect predictions incur large penalties, while appropriately
uncertain predictions are penalized less severely. This learning signal
encourages inference strategies that balance evidence aggregation with
uncertainty estimation, whereas sampling-based heuristics such as
ensembling reduce variance without modifying the underlying prediction
policy.

Relative to prior reinforcement learning with verifiable rewards,
Foresight Learning operates in open-world domains with sparse and
delayed feedback. Trajectory-level, group-relative optimization enables
stable credit assignment under long and variable horizons by comparing
alternative predictions generated under identical informational
constraints and evaluating them retrospectively after outcomes resolve.

This work has several limitations. Training is performed offline on
resolved events; while deployment-time feedback loops are under active
exploration, they are not evaluated in this study. Event specification
and outcome resolution rely on automated pipelines that may introduce
biases or coverage gaps, and the current experiments focus on binary
outcomes. Extending the framework to richer outcome spaces and fully
online settings remains an important direction for future work.

Overall, Foresight Learning demonstrates that effective supervision can
arise directly from chronologically evolving real-world data. By
incorporating outcome resolution into the training objective, the
framework points toward a broader role for outcome-based supervision in
extending verifiable reward-driven learning beyond closed-world tasks
and toward open-ended, real-world decision-making.

\section{Data and Model Availability}

The trained model, datasets, and data generation platform are publicly
available to support reproducibility and future research.

\noindent\textbf{Training data:}
\href{https://huggingface.co/datasets/LightningRodLabs/future-as-label-paper-training-dataset}
{Hugging Face dataset}

\noindent\textbf{Model weights:} 
\href{https://huggingface.co/LightningRodLabs/future-as-label-paper-step160}
{Hugging Face model repo}

\noindent\textbf{Data generation:} \url{https://lightningrod.ai/}

\bibliography{foresight}
\bibliographystyle{icml2026}

\end{document}